# A Language for Planning with Statistics

Nathaniel G. Martin and James F. Allen
Department of Computer Science
University of Rochester
Rochester, NY

## Abstract

When a planner must decide whether it has enough evidence to make a decision based on probability, it faces the *sample size problem*. Current planners using probabilities need not deal with this problem because they do not generate their probabilities from observations. This paper presents an event-based language in which the planner's probabilities are calculated from the binomial random variable generated by the observed ratio of one type of event to another. Such probabilities are subject to error, so the planner must introspect about their validity. Inferences about the probability of these events can be made using statistics. Inferences about the validity of the approximations can be made using interval estimation. Interval estimation allows the planner to avoid making choices that are only weakly supported by the planner's evidence.

## 1 INTRODUCTION

Planning relies on choosing the future actions most likely to be effective. Because actions are taken after they are planned, a planner's information is uncertain at planning time. Probabilities have been explored as a means of representing and reasoning about this uncertainty. In some domains, the necessary probabilities can be gathered by querying experts in the field. If such experts do not exist, an agent must be able to infer probabilities from observations. This paper develops a language that combines Allen's temporal interval reasoning [Allen, 1984] with statistical inference [Bickel and Doksum, 1977] to facilitate planning using inferences about probabilities.

If a planner must calculate its probabilities, it must decide when it has enough information to be confident of its calculations. Deciding when one is sufficiently confident of probabilities generated from observations is called the *sample size problem*. The sample size problem will be ubiquitous for planners that calculate probabilities from their experience. The most immediate incarnation of this problem is that of deciding whether choices are warranted by the evidence. A planner should make its decisions based on the probabilities about which it has good evidence, and discount the probabilities about which it is uncertain. One application of making decisions based on strength of evidence as well as probability is dealing with facts one is told. A planner may be told that a particular course of actions is better than another, but if its evidence is sufficiently strong, it may choose to ignore this information. The sample size problem also arises when reasoning about specifying actions to an intelligent reactive execution module [Martin and Allen, 1990b]. Here, the planner specifies details of its plans only if it is confident of the probabilities it has calculated. Yet another place the problem arises is in probabilistic solutions to the qualification problem [Martin and Allen, 1990a, Weber, 1989]. The planner adds as many qualifications as it can without making the event about which it is reasoning so specific that there are insufficient statistics to make necessary choices.

Feldman and Sproull [1977] deal with uncertainty by applying decision theory [Raiffa, 1970] to the problem of choosing appropriate actions guiding an A* algorithm. Horvitz [1988] uses decision theory to reason about partial results in planning. Johnson and Schubert [1982] use decision theory to control the cost of planning. More recently, Hartman [1990] has studied the same problem from a more formal perspective. All of these assume that probabilities are known beforehand. Moreover, none of these proposals includes an explicit representation of time.

Kanazawa and Dean [1989] propose a system that uses Bayes nets to make the computation of expected utilities sufficiently efficient to be used in a reactive execution architecture. They suggest using maximum entropy prior distributions as uninformative priors [Dean and Kanazawa, 1988]. Following Jaynes [1979], they choose the distribution that assumes one will receive the minimum amount of information from guessing. Kanazawa [1991] has recently coupled this Bayes net representation with a system for reasoning about probabilities and time.

Hanks [1988, 1990b, 1990a] has developed techniques for combining reasoning about time and probabilities. He is concerned with predicting future events given uncertain observations and actions. The planner then chooses the most effective actions given its beliefs about the state of the world in the future. Hanks also mentions maximum entropy as a guide to appropriate prior distributions.

Haddawy [1990, 1991] develops a formal logic of time and probability in which probability is represented by a modal operator over a temporal language. His system uses objective probabilities, a theory in which history determines chance in a fixed way. The theory of objective probabilities is a theory of causality that makes Bayesian inference valid.

Kanazawa and Hanks both choose a particular distribution from those warranted by their system's experience. The system maintains no information about the amount of evidence on which this distribution is based, so it cannot determine that, even though the probability of one prediction is higher than another, such a prediction rests on shaky foundations. Haddawy assumes an ontology in which probabilities are defined to be determined by the past. If it is used by a planning system, this ontology will rule out the possibility that its knowledge of the probabilities could be erroneous.

This paper explores interval estimation, a standard technique from statistical analysis, to deal with the sample size problem. It develops an event-based first-order language using temporal intervals and confidence intervals for reasoning about plans. Observations of instances of events are used to calculate confidence intervals with which the system represents and reasons about uncertainty. The language developed is similar to the event logic developed by Allen [1991] which is, in turn, based on the temporal logic described in Allen and Koomen [1983] and Allen [1984]. The confidence intervals are used in a manner similar to Kyburg's interval probabilities [Kyburg, 1983]. We provide an example of a planner making decisions based on the amount of evidence it has.

The proposal does not address the problem of generating beliefs from sensor input. It assumes that the beliefs have already been formed from the input, and the planner must decide what probability it should place on projections of its beliefs. We assume that the planner is buffered from the necessity of analyzing raw sensory information and generating control signals by an intelligent reactive system. The planner's task is to monitor the progress of the reactive system and give suggestions based on its observations.

Kaelbling [1990] investigates the possibility of applying interval estimation techniques to learning in embedded systems. She concludes that though learning algorithms that use interval estimation are slightly better than those that use point estimation, they are ill suited to learning in embedded systems because of their computational complexity and the difficulty of applying statistics to situations different from those in which they were gathered. Our use of interval estimation differs from Kaelbling's in that we apply the technique to a strategic planner that is buffered from its situation by its reactive execution module. Moreover, by reasoning about the best description of its current situation, the planner can apply statistics to situations different from those in which they were gathered.

**Example** As a running example, consider a robot engineer trying to couple two cars. This engineer has a program, **Old**, which it executes whenever it wants to couple cars. It has executed the program 1000 times, but has coupled successfully only 500 times. Recently, a new program, **New** was written and added to the engineer's repertoire. The old program remains an action the engineer can choose. Should the robot try the new program? A conservative guess of the probability of successfully coupling the cars using the new program might be 0.5, so the engineer might try it. If the engineer tries once and fails, it will believe the probability of coupling the cars using the new program is lower than the probability of success using the old program. The engineer could be less conservative and choose a higher prior for the new program, but what should that prior be? In general, how many times should the robot try the new program before concluding that the old program is better? □

## 2 KNOWLEDGE REPRESENTATION

This paper develops a first-order language that allows one to use statistics to reason about plans and actions. The language is concerned with five kinds of things: actions (a), event instances ($e_i$), temporal intervals ($t_i$), probability intervals ($i_i$), and $\alpha$-levels ($\alpha_i$). This language is similar to the language developed by Allen [1991]. It differs in the inclusion of probability intervals and $\alpha$-levels.

An event type is a set of event instances characterized by a sentence that constrains the temporal interval during which the instances of the event type occur. For simplicity, events in this paper are characterized only by their temporal properties. Therefore, a particular event instance can be specified by fixing the time at which the event occurs, $Time(e)$. We use the term *event* to represent an event type; event instances will be referred to as such. The characterization of an event is a sentence. We say that one event, $e_1$, subsumes another, $e_2$, if the characterization of $e_2$ is a logical consequence of the characterization of $e_1$.

Events express context and provide the basis for the calculation of the confidence intervals. Events encode the context of an action in its characterization. For example, being in the same city might characterize an event caused by any execution of the **Old** program. A particular event may also have other properties such as clear weather, but as long as the characterizing sentence is true, the event is said to hold. Probability is defined as the frequency of one event relative to another both in the past and in the future. For example, the probability that the engineer successfully couples two cars is the ratio of the number of successful attempts to the total number of attempts. Statistics about events change as the planner discovers more of the elements that make up the probability, but the probability itself does not change.

The temporal intervals associated with events allow the system to choose actions relative to an event, then order the events, allowing non-linear planning. Plans are generated as described by Allen [1991]. Temporal relations are specified using Allen's interval temporal logic [Allen, 1984]. These temporal intervals allow agents to reason about sequential and concurrent actions. For example, the robot engineer may need to reason that it must keep the coupler open while backing up if it wants to couple two cars.

Actions are the names of programs. When a program is executed, it causes an instance of an event. For example, a planner's program for coupling cars may simply back the engine until it hits a car. Clearly, if the train is on the wrong track, executing this program will not have the desired effect. The predicate Causes $(execute(\mathbf{a}, t), e)$ indicates that the program **a** was executed during time interval $t$ and caused event $e$. Some of the circumstances that affect the results of executing a program may be specified in the characterization of the event caused by the execution of that program; others may not. The circumstances that are specified in the characterization of such an event express preconditions of the event; those that are not make the events amenable to analysis by probabilities. That is, each event describes a set of event instances, each of which is different (at the very least in its time of occurrence). Ratios of the cardinalities of these sets are the probability of one event relative to another.

Statistics are maintained on the number of occurrences of an event, $occurrences(event\text{-}type)$. The reactive executor updates the planner's knowledge periodically. Each time a new temporal interval is added to the planner's knowledge base, it forward chains on this new information. If, in the process of forward chaining, it proves that an instance of event $e$ has occurred, it increments the value of $occurrences(event\text{-}type)$. As described here, the function $occurrences$ is not part of the language; instead, it is used to define the confidence intervals that are part of the language.

**Example** A single constraint characterizes an event such that an instance of it occurs whenever the agent attempts the **Old** action,

$$\forall(e)[\text{Causes }(execute(\mathbf{Old}, Time(e)), e) \Rightarrow \text{Old-Try }(e)].$$

This event represents all time intervals that match its characterization. A *Couple* event is characterized by,

$$\forall \ (c1, c2, t_1, t_2, e) \quad [\neg\text{Coupled }(c1, c2, t_1) \land$$
$$\text{Coupled }(c1, c2, t_2) \land$$
$$\text{Starts }(t_1, Time(e)) \land$$
$$\text{Ends }(t_2, Time(e))$$
$$\Rightarrow \quad \text{Couple }(c1, c2, e)].$$

That is, a couple event is one where before the event the cars were not coupled, and after the event the cars were coupled. We name the events characterized in this way by the characteristic predicate. That is

$$\forall(e)[\text{Couple }(c1, c2, e) \iff e \in Couple(c1, c2)]$$

is always true.

If we assume the planner has seen 1000 *Old-Try* event instances, of which 500 are also *Couple(c1,c2)* event instances, i.e.

$$occurrences(Old\text{-}Try) = 1000$$
$$occurrences(Couple(c1, c2) \cap Old\text{-}Try) = 500$$

then we have characterize the situation of the robot engineer at the beginning of the example mentioned above. The event described by $Couple(c1, c2) \cap Old\text{-}Try$ is the least constrained one that subsumes both $Couple(c1, c2)$ and $Old\text{-}Try$.

When the robot is given the new program for coupling cars, it will need to be able to distinguish the event in which it tries this program. This event will be called *New-Try*,

$$\forall \ (e) \quad [\text{Causes }(execute(\mathbf{New}, Time(e)), e)$$
$$\Rightarrow \quad \text{New-Try }(e)].$$

□

The $\alpha$-levels are the probability that the parameter being estimated falls outside the confidence interval computed. Confidence intervals are calculated to be of the sizes of the system's $\alpha$-levels. An $\alpha$-level confidence interval for a parameter $p$ is a random interval for which the probability that the interval contains $p$ is $\alpha$. Confidence intervals represent the strongest possible constraints on the location of the parameter given the data observed.

## 3 INFERENCE

Using statistics on events, the planner can compute constraints on the probability distributions consistent with its knowledge. We define $PCA_{1-\alpha}(e_g, e_a)$ to be the $1-\alpha\%$ confidence interval for the mean of the binomial random variable calculated from the occurrences of two events, $e_g$ and $e_a$. We call the event in which the action is tried the reference event, the event after which the goal holds the success event.

Such a confidence interval can be approximated using the DeMoivre-Laplace theorem on the approximation of binomial distributions by normal distributions. Given that $z_{\alpha/2}$ is the portion of the standard normal distribution such that

$$P[Z > -z_{\alpha/2}] = P[Z < z_{\alpha/2}] = \alpha/2$$

the confidence interval for n Bernoulli trials with y successes is approximated by:

$$\left[ \frac{y + \frac{z_{\alpha/2}^2}{2} - z_{\alpha/2}\sqrt{\frac{y(n-y)}{n} + \frac{z_{\alpha/2}^2}{4}}}{n + z_{\alpha/2}^2}, \frac{y + \frac{z_{\alpha/2}^2}{2} + z_{\alpha/2}\sqrt{\frac{y(n-y)}{n} + \frac{z_{\alpha/2}^2}{4}}}{n + z_{\alpha/2}^2} \right]$$

To use these confidence intervals one must assume that the normal distribution is a good approximation to the binomial. This will be the case when the smaller of $np$, $n(1-p)$ is less than five, where $p$ is the parameter for the binomial distribution being approximated.

Where there are only a few instances of an event, exact bounds can be computed. The cost of computing these bounds is high, but they can be precomputed and stored in a relatively small table for those cases in which the normal approximation is invalid. The exact $\alpha$-level confidence interval for the parameter of a binomial random variable generated by $n$ Bernoulli trials with $y$ successes ($y > 0$) will be $[p_l, p_u]$, where $p_l$ is the unique solution to the equation,

$$\sum_{i=y}^{n} \binom{n}{i} p_l^i (1-p_l)^{n-i} = \alpha,$$

and $p_u$ is the unique solution to the equation,

$$\sum_{i=0}^{y} \binom{n}{i} p_u^i (1-p_u)^{n-i} = \alpha.$$

Tables of confidence intervals for binomial random variables can be found in [Clopper and Pearson, 1934] and [Fisher and Yates, 1963]. There is a discussion of the trade-off between the exact confidence interval and the approximation in [Kendall and Stuart, 1961].

To calculate confidence intervals, the planner needs to know the number of occurrences, $n$, of instances of the reference event and the number of occurrences, $y$, of instances of the event subsumed by the reference event that are also subsumed by the success event. That is, the planner will need to know $n = occurrences(e_a)$ and $y = occurrences(e_a \cap e_g)$.

Confidence intervals are represented in the language by constants allowing sentences about the systems constraints on probabilities. One interval is "$\prec$" another if the upper bound of the first is less than the lower bound of the second. The intervals are equal if both bounds correspond. Intervals that overlap are said to be incomparable. That is, $[0.5, 0.6] \prec [0.7, 0.9]$ but $\neg([0.5, 0.6] \prec [0.5, 0.9])$ and $\neg([0.5, 0.9] \prec [0.5, 0.6])$. The planner chooses the action whose confidence interval at a given $\alpha$-level was the highest among all applicable actions using $\prec$.

A predicate describing the planner's preferred action, Best $(\mathbf{a}, e_g, \alpha)$, can be defined using this language. The planner prefers an action $\mathbf{a}$ if and only if the statistics it has about the event in which the action occurred give clear indication that $\mathbf{a}$ is most likely to cause an event that leads to the goal (i.e., Goal(e)). The predicate can be defined by the following conditional:

(1) $\forall (e_g, e_a, \alpha, \mathbf{a})$ [Goal $(e_g) \wedge$
　　　Causes $(execute(\mathbf{a}, Time(e_a)), e_a) \wedge$
　$\forall (e_b, \mathbf{b})$ [$\mathbf{a} \neq \mathbf{b} \wedge$
　　　Causes $(execute(\mathbf{b}, Time(e_b)), e_b) \wedge$
　　　$PCA_{1-\alpha}(e_g, e_b) \prec PCA_{1-\alpha}(e_g, e_a)$]
　$\Rightarrow$ Best $(\mathbf{a}, e_g, \alpha)$]

This predicate says that action $\mathbf{a}$ is best if and only if the planner's knowledge constrains the probability of its success to be higher than any other action.

**Example** Suppose the number of occurrences of *New-Try* and *Couple(c1,c2)* are as follows:

$occurrences(New\text{-}Try) = 2$
$occurrences(New\text{-}Try \cap Couple(c1, c2)) = 1.$

In this example the engineer uses only a .05 $\alpha$-level to generate its probability constraints. From the number of occurrences of the preceding events, the following constraints on the probability of the successful execution of the actions can be generated:

$PCA_{.95}(Couple(c1, c2), Old\text{-}Try) = [0.4691, 0.5309]$
$PCA_{.95}(Couple(c1, c2), New\text{-}Try) = [0.0254, 0.9747].$

The planner knows that if these intervals[1] do not overlap it can choose the higher interval, and with probability at least $1 - \alpha$, this decision is correct. This information is encoded in (1). Here, the planner can prove neither

Best $(Couple(c1, c2), Try\text{-}Old, .05, \mathbf{Old})$

nor

Best $(Couple(c1, c2), New\text{-}Try, .05, \mathbf{New})$

using (1). It must give up. $\square$

---

[1] The confidence interval for **New** is exact.

When intervals overlap, the statistics do not indicate a clear choice at the $\alpha = .05$ confidence level. When the statistics do not provide clear guidance, the robot might fall back on heuristics. One source of these heuristics might be suggestions by the programmer writing the programs about which the planner reasons.

**Example** The programmer might tell the robot that the new couple program is better than the old one, since, presumably, this was the reason for writing it. The planner should take this advice only if it does not conflict with its experience. It might therefore translate this advice into a rule saying it should choose the new program only as long as there is no clear evidence that this program is inferior to the old one. Such a rule can be defined by the following conditional:

(2) $\forall(e)$ [ $\neg(PCA_{.95}(Couple(c1, c2), Old\text{-}Try) \prec$
$PCA_{.95}(Couple(c1, c2), New\text{-}Try)) \wedge$
$\neg(PCA_{.95}(Couple(c1, c2), New\text{-}Try) \prec$
$PCA_{.95}(Couple(c1, c2), Old\text{-}Try)) \Rightarrow$
Best $(Coupled, e, .05$ **New**$)$ ].

This sentence states that whenever there is insufficient information to choose between the alternatives it should choose the **New** program. Using it, planner can prove Best $(Couple(c1, c2), e, .05, \textbf{New})$. □

When the planner has clear evidence that one action is better than another, it need not rely on heuristics. This will be the case when evidence about the effectiveness of the **New** action overwhelms evidence about the effectiveness of the **Old** action. More importantly, the planner will cease to use the **New** action if it gets clear information that this action does not result in improved performance.

**Example** Suppose, after applying the default rule 100 times, the robot finds that **New** has resulted in cars being coupled 70 times. The number of times instances of the *New-Try* event and the *Couple(c1,c2)* event have occurred are

$occurrences(New\text{-}Try) = 100$
$occurrences(Couple(c1, c2)) = 70$,

generating new constraints on probability,

$PCA_{.95}(Couple(c1, c2), New\text{-}Try) = [0.6041, 0.7811]$.

Because $[0.4691, 0.5309] \prec [0.6041, 0.7811]$, the robot chooses the **New** program and will continue to do so unless it discovers that the probability constraints for the **New** program fall below the probability constraints for the **Old** program.

If after the initial two executions of the **New** program, the next seven cause *Couple(c1,c2)* events, the robot stops relying on the heuristic and begins relying on its own experience. The exact confidence intervals for eight successes in nine trials is $[0.5709, 0.9944]$ whereas the exact confidence interval for seven successes in eight trials is $[0.5294, 0.9937]$. Therefore, seven immediate successes (i.e. the robot has seen eight successes in all) are sufficient to make the robot choose the **New** program without using the heuristic.

Suppose, alternatively, that the robot discovers that it has successfully coupled cars only 30 times after applying the default rule 100 times. Now the number of occurrences of the event in which it tried the **New** program and the event in which the cars were coupled are

$occurrences(New\text{-}Try) = 100$
$occurrences(Couple(c1, c2)) = 30$

generating new constraints on probability,

$PCA_{.95}(Couple(c1, c2), \text{New-Try}) = [0.2189, 0.3959]$.

The planner chooses to return the **Old** program because $[0.2189, 0.3959] \prec [0.4691, 0.5309]$.

If after the initial two observations of the *Try-New* event, the next eight event instances are not also instances of the *Couple(c1,c2)* event the robot rejects the heuristic preference for the **New** program. Because $[0.0057, 0.4292]$ is the exact confidence intervals for one success in nine trials and $[0.0064, 0.4707]$ is the interval for one success in eight trials, the robot will need at least eight immediate failures to choose the **Old** action against the advice of the programmer. □

The robot makes a choice when it has enough information to do so; it may reason further or rely on heuristics when it cannot. As the robot gathers more information, it can make more choices based on its information about the probability of success of actions and rely less on guesses.

## 4 PLANNING

Besides choosing actions, a planner must deal with preconditions and composite actions. Preconditions are important because some details may dramatically affect the probability of success of the action chosen. The planner must be able to take these details into account. The planner cannot take into account everything it knows about the current situation because, in part, there will be only one occurrence of such an event. This is the sample size problem. To solve the problem, the planner chooses the most constrained event that subsumes the current situation for which it has sufficient statistics to make a choice. We call this event the initial event.

To facilitate these solutions, a new event, $e_p$, which represents the context of the action, is added to the computation of the probability of the goal given the action. The planner computes $PCA_{1-\alpha}(e_g, e_a, e_p)$ from $n = occurrences(e_a \cap e_p)$ and $y = occurrences(e_a \cap e_g \cap e_p)$. The planner chooses the preconditions that produce the highest comparable confidence interval.

**Example** The engineer is more likely to successfully couple cars if the cars and the engine are in the same city. These constraints can be added to the event against which the success of the action is to be measured. For example, suppose we have a new event called a *Pre-Try* in which the cars and the engine are in the same city. This event will be characterized by:

$$\forall\ (e, c1, c2, city) \quad [\text{In } (c1, city, Time(e)) \land$$
$$\text{In } (c2, city, Time(e)) \land$$
$$\text{In } (Me, city, Time(e))$$
$$\Rightarrow \quad \text{Pre-Try}(e)\ ].$$

Another event, *Any*, describes a situation with no constraints in its characterization.

$$\forall(e)[Any(e)]$$

Since every event that subsumes *Old-Try* ∩ *Any* also subsumes *Old-Try* ∩ *Pre-Try*, the planner will have more evidence for *Old-Try* ∩ *Any*. Since, however, success is unlikely for *Old-Tries* that were not also subsumed by *Pre-Try*, the probability of success will be higher for *Old-Try* ∩ *Pre-Try*.

To choose the appropriate preconditions, the planner will also need to know that both **Old** and **New** are programs whose intention is to couple cars. This can be indicated by generating a new event that is subsumed by either *Try-Old* or *Try-New*, i.e.,

$$\forall(e)[\text{Old-Try }(e) \lor \text{New-Try }(e) \Rightarrow \text{Try }(e)].$$

Suppose that the statistics mentioned above have no particular context and that there are 800 instances of an event that subsumes *Try* ∩ *Pre-Try*. Suppose also that the number of occurrences of the success event described above is the same as the number of occurrences of success for *Try* ∩ *Pre-Try*. That is,

$$occurrences(Try \cap Pre\text{-}Try) = 800$$
$$occurrences(Couple(c1, c2) \cap Try \cap Pre\text{-}Try) = 501$$
$$occurrences(Try \cap Any) = 1002$$
$$occurrences(Couple(c1, c2) \cap Try \cap Any) = 501.$$

These statistics lead to the following probability constraints:

$$PCA_{.95}(Couple(c1, c2), Try, Pre\text{-}Try) =$$
$$[0.5909, 0.6579]$$
$$PCA_{.95}(Couple(c1, c2), Try, Any) =$$
$$[0.4691, 0.5309].$$

Since $[0.4691, 0.5309] \prec [0.5922, 0.6591]$, the planner chooses *Pre-Try* as the preconditions to the action. □

The planner ignores preconditions about which it has insufficient information. Even though they may affect the probability of the goal, they can be ignored with relative safety because they occur infrequently. Choosing preconditions in this manner is similar to assuming preconditions as suggested by Allen [1991].

**Example** Suppose the engineer can recognize when the cars loaded or empty.

$$\forall\ (e, c1, c2, city) \quad [\text{Loaded } (c1, Time(e)) \land$$
$$\text{Empty } (c2, Time(e)) \land$$
$$\text{In } (c1, city, Time(e)) \land$$
$$\text{In } (c2, city, Time(e)) \land$$
$$\text{In } (Me, city, Time(e))$$
$$\Rightarrow \quad \text{Pre-Try2 }(e)]$$

The ability to recognize the state of the cars may be important if the engineer's task is to move cargo. Due to the large number of such events, however, the engineer may have weak statistics on them.

Suppose that there are 100 instances of an event that subsumes *Try* ∩ *Pre-Try2*. Suppose also that 75 of the instances of the success events described above are are instances of *Try* ∩ *Pre-Try2*. That is,

$$occurrences(Try \cap Pre\text{-}Try2) = 100$$
$$occurrences(Couple(c1, c2) \cap Try \cap Pre\text{-}Try) = 75$$

These statistics lead to the following probability constraints:

$$PCA_{.95}(Couple(c1, c2), Try, Pre\text{-}Try2) =$$
$$[0.6570, 0.8245].$$

Since neither $[0.6570, 0.8245] \prec [0.5922, 0.6591]$ nor $[0.5922, 0.6591] \prec [0.6570, 0.8245]$, the planner chooses *Pre-Try* as the preconditions to the action again. In this case it chooses to ignore a precondition because it does not have enough information about success relative to the precondition. As far as the planner can tell from the statistics, success assuming one event is the same as success assuming the other. □

Once the planner has chosen the appropriate preconditions for its actions, it chooses actions relative to these preconditions as outlined above.

The planner must also deal with sequences of actions. Due to space restrictions, there is only room for a cursory overview of the details of generating such sequences.

When choosing an action in a sequence, the planner chooses relative to a hypothetical event caused by executing the actions chosen earlier. For example, in choosing the second action of a two-action plan $(A_1, A_2)$, it should select the second action in the context of the event caused by the execution of $A_1$ in the initial event. Since action $A_1$ may have many possible effects, this new event may be no simpler than was the complete description of the current situation. The planner simplifies this event by reasoning relative to an event for which it has sufficient statistics to choose an appropriate action for the second step of the plan and which subsumes the event caused by executing $A_1$.

When selecting an event from which to choose subsequent actions, the planner must first recognize that no single action is adequate. Because the effects of actions are uncertain, one possible result of any action is that the goal will hold. As a heuristic the planner might assume that no single action effectively achieves the goal when assuming it performs any single action makes the goal no more likely than assuming it does nothing. Here the planner can be confident that by using time to continue planning, it will miss deadlines. If the goal is part of the current situation, doing nothing is most likely to cause an event in which the goal holds than doing nothing, as it causes an event that subsumes the maximum number of events and is therefore most likely to subsume the current situation.

Once it has realized that it must generate a series of sub-goals to achieve its main goal, it can then deal with each sub-goal as a separate problem. The problems are not really separate, however, because choosing actions that achieve remaining sub-goals may reduce the probability that actions already chosen achieve their sub-goals. The planner avoids such interaction by choosing remaining actions relative to a hypothetical event that subsumes the event caused by executing actions chosen earlier. The order in which the planner chooses actions is unimportant because the temporal logic allows both constraints that precede and constraints that follow the execution of actions.

If the planner has sufficient statistics to reason about sequences of events, it will use them. That is, if it can actually make subsequent choices given the desired results of previous choices, the planner will make the choices. In situations requiring planning, it is unlikely that the planner will have good statistics for long sequences of actions, however. Except for those sequences that are chosen frequently, the statistics are likely to be very weak for the choices the planner must make in long plans. Note that it is the small sample size, not low probability, that makes such decisions untenable. The planner may have actually succeeded every time it chose an action in a very constrained event; it just has not made those choices often enough to be confident in them.

If the planner has insufficient statistics to make subsequent choices, it may rely on heuristics like the one described by formula (2). For example, a good heuristic would be to wait until further information arrives. If a planner has a partial plan it cannot complete, it might simply specify that partial plan to the reactive execution system and hope for the best. Even if the partial plan is insufficient to actually achieve the goal, the planner may have more information when it needs to replan.

If the planner has no applicable heuristics or world knowledge, it will assume the actions are independent. Such an assumption may be incorrect, but a planner that uses statistics will at least have evidence that the plans it is generating are ineffective when the statistics begin to reflect its current strategy's low probability of success. For example, if the planner cannot recognize the event in which the cars and the engine are in the same city, it will continue to try to couple the cars, but will succeed only when the unrecognizable precondition holds. If the engine and cars are rarely in the same city, the probability of success for the engineer's couple programs will fall, and the engineer's confidence in this low probability will increase. Eventually, the planner will become confident enough that the action rarely succeeds that it will stop attempting it.

## 5 CONCLUSION

A language for reasoning with statistics gives planners the ability to reason about the strength of their evidence. By reasoning about the strength of its evidence, a planner can discount weak evidence as a reason for preferring one action over another. As far as we are aware, no other formalism combines temporal reasoning and reasoning about evidence. Systems that gather information and generate plans based on that information will need this ability.

A shortcoming of the proposal as presented here is the weakness of the statistical tests used. Generating confidence intervals and comparing them is wasteful of the planner's valuable data. We are studying other statistical tests that make better use of the data. Another problem is choosing preconditions based on estimations of the probability of the goal given the preconditions. A better criterion is the information the preconditions provides for the the choice of actions. Measures of information may perform better than do constraints on probability.

An unnecessary restriction of the presentation is its adherence to the frequentist view of probabilities. All of the techniques presented in this paper are equally valid if one uses Bayesian interval estimation rather than confidence intervals. Indeed, even many of the more powerful statistical tests under study have Bayesian correlates.

A shortcoming this proposal shares with others is the large number of events needed for general purpose planning. In this system, events play the part of operators in STRIPS. Still, probabilities may suggest a solution to this problem. One could control the number of events the planner needs to consider by ensuring that the event occurs frequently. If such an assurance can be made, the planner can assume that if it has no statistics about a particular event, that event is rare. Such assurances may be possible if events are generated through cluster analysis. If the events are generated in this way, the planner may safely assume that only rare events will have no statistics.

Acknowledgments

We would like to thank George Ferguson for his many insightful commments on this work. We would also like to thank Steve Hanks and Bulent Murtezaoglu for comments on earlier drafts of the paper. This material is based on work supported by ONR/DARPA under grant number N0014-82-K-0193 and under AF-Rome Air Development Center contract number F30602-92-C-0010.

References

[Allen and Koomen, 1983] James Allen and Johannes Koomen. Planning using a temporal world model. In *IJCAI-83*, pages 741–747, 1983.

[Allen, 1984] James F. Allen. Towards a general theory of action and time. *Artificial Intelligence*, 23(2):123–145, 1984.

[Allen, 1991] James F. Allen. Planning as temporal reasoning. In *KR-91*, pages 3–14, 1991.

[Bickel and Doksum, 1977] Peter J. Bickel and Kjell A. Doksum. *Mathematical Statistics: Basic Ideas and Selected Topics*. Holden-Day, Inc., Oakland, CA, 1977.

[Clopper and Pearson, 1934] C. J. Clopper and E. S. Pearson. The use of confidence or fiducial limits illustrated in the case of the binomial. *Biometrika*, 26:404–413, 1934.

[Dean and Kanazawa, 1988] Thomas Dean and Keiji Kanazawa. Probablistic temporal reasoning. In *AAAI-88*, pages 125–132, 1988.

[Feldman and Sproull, 1977] Jerome Feldman and Robert Sproull. Decision theory and artificial intelligence II: The hungry monkey. *Cognitive Science*, 1:158–192, 1977.

[Fisher and Yates, 1963] R. A. Fisher and F. Yates. *Statistical Tables for Biological Agricultural and Medical Research (6th ed.)*. oliver and Boyde, Edinburg, 1963.

[Haddawy, 1990] Peter Haddawy. Time, chance, and action. In *Uncertainty in AI 90*, pages 147–153, 1990.

[Haddawy, 1991] Peter Haddawy. A temporal probability logic for representing actions. In *KR-91*, pages 313–324, 1991.

[Hanks, 1988] Steve Hanks. Representing and computing temporally scoped beliefs. In *AAAI-88*, pages 501–505, 1988.

[Hanks, 1990a] Steve Hanks. Practical temoral projection. In *AAAI-90*, pages 158–163, 1990.

[Hanks, 1990b] Steven John Hanks. *Projecting Plans for Uncertain Worlds*. PhD thesis, Yale University, New Haven, CT, 1990.

[Hartman, 1990] Leo B. Hartman. *Decision Theory and the Cost of Planning*. PhD thesis, University of Rochester, Rochester, NY 14627, 1990.

[Horvitz, 1988] Eric J. Horvitz. Reasoning under varying and uncertain resource constraints. In *AAAI-88*, volume 1, pages 111–116, 1988.

[Jaynes., 1979] E. T. Jaynes. Where do we stand on maximum entropy? In R. D. Levine and M. Tribus, editors, *The Maximum Entropy Formalism*, pages 279–293. MIT Press, 1979.

[Johnson and Schubert, 1982] D. T. Johnson and Lenhart K. Schubert. A planning control strategy that allows for the cost of planning. In *Proc. 6th Eur. Meet. on Cybernetics and Sys. Research*, pages 1–7, 1982.

[Kaelbling, 1990] Leslie Pack Kaelbling. *Learning in Embedded Systems*. PhD thesis, Stanford University, Stanford, CA, 1990.

[Kanazawa and Dean, 1989] Keiji Kanazawa and Thomas Dean. A model for projection and action. In *IJCAI-89*, pages 985–990, 1989.

[Kanazawa, 1991] Kaiji Kanazawa. A logic and time nets for probabilistic inference. In *AAAI-91*, 1991.

[Kendall and Stuart, 1961] M. G. Kendall and A. Stuart. *The Advanced Theory of Statistics Vol II (2nd ed)*. Hafner Publishing Co., New York, 1961.

[Kyburg, 1983] Henry E. Kyburg, Jr. The reference class. *Philosophy of Science*, 50:374–397, 1983.

[Martin and Allen, 1990a] Nathaniel G. Martin and James F. Allen. Abstraction in planning: A probabilistic approach. Presented at the Workshop on Automatic Generation of Approximations and Abstractions, 1990.

[Martin and Allen, 1990b] Nathaniel G. Martin and James F. Allen. Combining reactive and strategic planning through decomposition abstraction. In *Workshop on Innovative Approaches to Planning, Scheduling and Control*, pages 137–143, 1990.

[Raiffa, 1970] Howard Raiffa. *Decision Analysis: Introductory Lectures on Choices under Uncertainty*. Addison-Wesley, Reading, MA, 1970.

[Weber, 1989] Jay C. Weber. A parallel algorithm for statistical belief refinement and its use in causal reasoning. In *IJCAI-89*, August 1989.